\documentclass[sn-mathphys]{sn-jnl}

\usepackage{amsmath,graphicx}
\usepackage{multirow}
\usepackage{hyperref}
\usepackage[misc]{ifsym} 
\usepackage{color}
\usepackage{subcaption}
\usepackage{sidecap}
\usepackage{amsfonts}
\usepackage{comment}
\usepackage{adjustbox}
\usepackage[T1]{fontenc}

\usepackage{graphicx}

\jyear{2021}%

\theoremstyle{thmstyleone}%
\theoremstyle{thmstyletwo}%

\theoremstyle{thmstylethree}%

\begin{document}

\title[Pose-based Tremor Type and Level Analysis for PD from Video]{Pose-based Tremor Type and Level Analysis for Parkinson's Disease from Video    \vspace{-0.3cm} }

\author[1]{\fnm{Haozheng} \sur{Zhang}}
\author[2]{\fnm{Edmond S. L. } \sur{Ho}}
\author[1]{\fnm{Xiatian} \sur{Zhang}}
\author[3,4]{\fnm{Silvia} \sur{Del Din}}
\author*[1]{\fnm{Hubert P. H.} \sur{Shum}}\email{hubert.shum@durham.ac.uk\vspace{-0.5cm}}

\affil[1]{\orgdiv{Department of Computer Science}, \orgname{Durham University}, 
\orgaddress{\city{Durham},\country{UK}}}

\affil[2]{\orgdiv{School of Computer Science}, \orgname{University of Glasgow}, \orgaddress{\city{Glasgow},\country{UK}}}

\affil[3]{\orgdiv{Faculty of Medical Sciences} \orgname{Newcastle University}, \orgaddress{\city{Newcastle upon Tyne},  \country{UK}}}

\affil[4]{\orgdiv{National Institute for Health and Care Research Newcastle Biomedical Research Centre}, \orgname{The Newcastle upon Tyne Hospitals NHS Foundation Trust}, \orgaddress{\city{Newcastle upon Tyne},  \country{UK}\vspace{-0.6cm}}}



%
\abstract{
\textbf{Purpose}:
Current methods for diagnosis of PD rely on clinical examination. The accuracy of diagnosis ranges between 73\% and 84\%, and is influenced by the experience of the clinical assessor. Hence, an automatic, effective and interpretable supporting system for PD symptom identification would support clinicians in making more robust PD diagnostic decisions.

\textbf{Methods}:
We propose to analyze Parkinson's tremor (PT) to support the analysis of PD, since PT is one of the most typical symptoms of PD with broad generalizability. To realize the idea, we present SPA-PTA, a deep learning-based PT classification and severity estimation system that takes consumer-grade videos of front-facing humans as input. The core of the system is a novel attention module with a lightweight pyramidal channel-squeezing-fusion architecture that effectively extracts relevant PT information and filters noise. It enhances modeling performance while improving system interpretability.

\textbf{Results}:
We validate our system via individual-based leave-one-out cross-validation on two tasks: the PT classification task and the tremor severity rating estimation task. Our system presents a 91.3\% accuracy and 80.0\% F1-score in classifying PT with non-PT class, while providing a 76.4\% accuracy and 76.7\% F1-score in more complex multiclass tremor rating classification task.

\textbf{Conclusion}:
Our system offers a cost-effective PT classification and tremor severity estimation results as warning signs of PD for undiagnosed patients with PT symptoms. In addition, it provides a potential solution for supporting PD diagnosis in regions with limited clinical resources.}

\vspace{-3mm}
\maketitle     

\vspace{-7mm}
\section{Introduction}
\vspace{-1mm}
\normalsize Parkinson's disease (PD) is the second most common progressive neurological disorder, affecting an estimated 10 million people globally~\cite{Chopade2023}. It is characterized by the loss of dopaminergic neurons within the substantia nigra region of the brain, resulting in motor dysfunction \cite{mhyre2012parkinson}. Existing PD diagnosis is mainly based on the clinical assessment of PD symptoms, medical history, l-dopa, and dopamine responses \cite{mostafa2019examining}. The clinical diagnostic accuracy is approximately 73\%-84\%~\cite{rizzo2016accuracy}, and may be affected by medical experts' subjective opinions and experiences. An automatic, efficient, and interpretable PD assessment system would support clinicians in making more robust diagnostic decisions.

Recent research in PD diagnosis with machine learning using human-centric visual, audio, and movement features has shown promising results.
Models based on neuroimaging~\cite{zhang2020survey} and cerebrospinal fluid biomarkers~\cite{wang2020early} provide an accurate diagnosis but are costly and intrusive, making them unsuitable for large-scale pre-diagnosis. Non-intrusive methods with speech~\cite{vasquez2018multimodal} are limited by their generalizability due to the significant difference in language and pronunciation for patients from different geographical areas. 
Although gait disturbance is not typically the primary symptom of early-onset PD \cite{hausdorff2009gait,rizek2016update}, over 70\% of these patients exhibit at least one form of tremor~\cite{rizek2016update}. Hence, identifying Parkinson's Tremor (PT) is seen as a more generalizable approach for assisting in early PD diagnosis. To date, hand tremors-based studies mostly rely on wearable sensor data~\cite{hssayeni2019wearable}. However, the use and set-up of wearable technology may be time and resource-consuming~\cite{hssayeni2019wearable}. Video-based analysis with consumer-grade cameras is preferable as a more cost-effective solution without disrupting the natural behavior of the participants. 

We propose a novel open-source\footnote{\href{https://github.com/zhz95/SPA-TPA}{ https://github.com/zhz95/SPA-TPA}} video-based deep learning system for PT classification and tremor severity estimation to assist the pre-diagnosis of PD with PT symptoms. We first extract the upper body human pose from videos as an effective feature for tremor analysis. We then design a graph neural network with a novel Pyramidal Channel-Squeezing-Fusion (PCSF) architecture that learns the attention by representing the joint-wise relevancy in a hierarchical manner. Such attention values allow interpretation of the features considered by the network for decision-making. Our solution outperforms existing ones in PT analysis, achieving 91.3\% accuracy and 80.0\% F1-score in PT classification, 76.4\% accuracy and 76.7\% F1-score in tremor rating classification. 

Compared with our preliminary work \cite{zhang2022miccai} that only focuses on tremor-type classification, we have the following technical improvements: (1) Adapting the system for tremor rating estimation. (2) Supplementing our system with the Eulerian video magnification to enhance the subtle tremors for better feature extraction. (3) Adding an examination with the Nyquist limits to test whether the input videos are suitable for tremor analysis. (4) Improving pose extraction by employing the state-of-the-art AlphaPose algorithm and conducting comprehensive experiments to evaluate its performance improvement. (5) Evaluating our system via a more challenging individual-based leave-one-out cross-validation to improve system robustness. (6) Conducting extra experiments with ablation studies and visualizations. 

\vspace{-3mm}
\section{Method}
\begin{figure*}
\includegraphics[width=\textwidth]{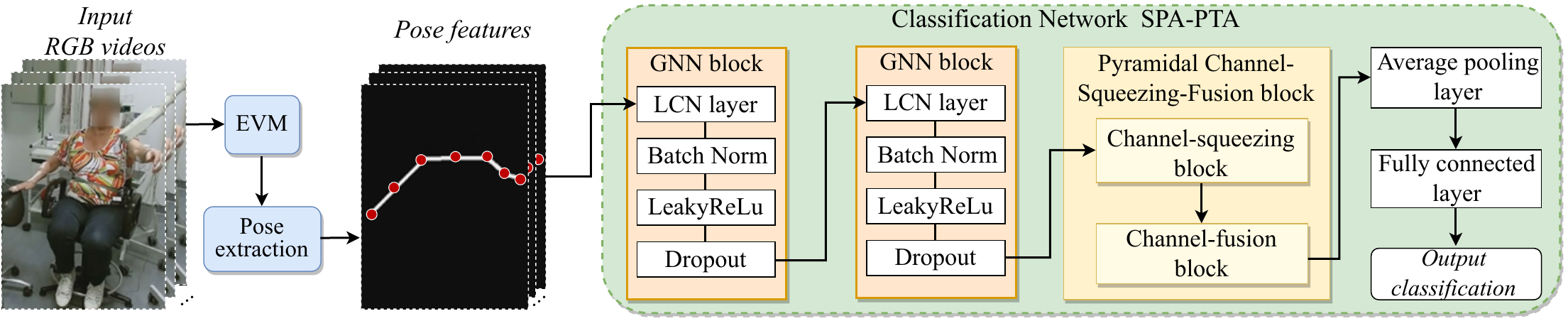}
\caption{The framework of our system: we use EVM to enhance the subtle tremors in the original videos, then pass videos to the pose extraction process. We classify the extracted pose features by SPA-PTA with a novel PCSF design.}
\vspace{-6mm}
\label{net}
\end{figure*}

\vspace{-1mm}
Fig.~\ref{net} shows the overview of our system. Its input is a set of videos showcasing a patient sitting in an upright posture, performing various actions such as keeping arms parallel to the ground. The human joint position features are extracted from the videos using AlphaPose~\cite{alphapose}, a state-of-the-art pose-estimation algorithm. These features are then fed into the Spatial Pyramidal Attention network for PT type and level Analysis (SPA-PTA).

\vspace{-3mm}
\subsection{Eulerian Video Magnification}
\vspace{-1mm}
We employ Eulerian Video Magnification (EVM) as a signal processing method ~\cite{Liu2023EVM} to enhance the subtle tremors and reduce noise and artifacts in the videos. This is motivated by previous research findings~\cite{zhang2022miccai} that deep neural network models paid more attention to human wrists during PT classification, indicating that magnifying subtle hand and wrist motions 
can be beneficial for tremor feature learning. Before applying EVM, we checked the Nyquist limits~\cite{Condon2016limit} to examine whether our video frequency is valid for tremor analysis. Specifically, the video frame rate should be at least twice the highest frequency of tremor motions. 
As existing research~\cite{Delval2016HZ} has shown that PT typically occurs between $3$ and $7Hz$, our video with $30Hz$ fulfils the requirement.

\vspace{-3mm}
\subsection{Pose Extraction}
\vspace{-1mm}
We extract the 2D pose features from the EVM-processed videos by 
AlphaPose~\cite{alphapose}. Compared to previous work using OpenPose~\cite{zhang2022miccai}, AlphaPose is superior as it demonstrates 25\% improved pose estimation performance on average precision and average recall metrics in multiple datasets. We prefer 2D poses to 3D ones, as current 3D pose estimation techniques are less mature, and they generally introduce noise particularity in the depth dimension~\cite{wang2020motion}, making them less suitable for sensitive features like tremors. We use AlphaPose to estimate 17 COCO-format~\cite{alphapose} body keypoints and extract $(x,y,c)$ features, where $(x,y)$ represent the 2D coordinate and $c$ is a confidence score that reflects the estimation accuracy. Consistent with previous work~\cite{zhang2022miccai}, we utilize the top half of the body keypoints (shown in Fig.~\ref{fig5}) for PT classification. It disregards less relevant lower-body features to enhance model efficiency and reduce potential bias because of the observation that PT generally occurs on the upper body, specifically on the hands and arms~\cite{sveinbjornsdottir2016clinical}. In addition, we omit the head joints as the participants' faces are generally obscured in medical videos to preserve their privacy. Furthermore, we normalize the pose to mitigate bias resulting from inherent video differences. In order to mitigate global translations in the pose, we align the mean location of the neck and two hip joints as the global origin. Subsequently, all joint positions are expressed as relative values to this established origin.

\vspace{-3mm}
\subsection{Classification Network}
\vspace{-1mm}
We propose the SPA-PTA for PT analysis by the PT classification task and an extended tremor severity estimation task. SPA-PTA is composed of two graph neural network (GNN) blocks with a spatial attention mechanism, along with a novel pyramidal channel-squeezing-fusion block designed to learn the joint-wise relevancy. 
\\
\noindent \textbf{GNN Block with Spatial Attention Mechanism:}
\\
We consider using Graph Neural Networks (GNN) for PT analysis, which are effective in modeling relational data, unlike images that are in a grid structure. In particular, human poses can be considered as a relational graph structure \begin{math} G = (V,E)\end{math}~\cite{yan2018spatial}, with the nodes representing the joints and the edges representing the skeletal structure across time.   

Formally, \begin{math} \{V = {v_{pq}}\}\end{math} represents the set of joint positions, where \begin{math} v_{pq} \end{math} is the \textit{p}-th 
joint at the \begin{math} q\end{math}-th frame. The set of edges, \textit{E}, consists of (i) spatial edges connecting different joints in space, and (ii) temporal edges connecting the same joint across consecutive frames.

We propose a spatial attention mechanism to enhance the performance of classification and improve the interpretability of our system. Specifically, it helps interpret the significant joints that the network identifies during classification by computing the attention weight of each joint per frame and its temporal aggregation. Moreover, it allows the system to learn the attention of the target joint by considering its relevancy with other joints. The fundamental expression is as $\mathbf{h_{i}}= \sigma \left( \sum_{j\in \mathcal{N}^{i}} \mathbf{W}_{j}^{i} \mathbf{x}_j \hat{a}_{ij} \right)$ 
where $\sigma$ is an activation function, $\mathbf{W}_{j}^{i}$ is the learnable attention weight between the target node \textit{i} and the related node \textit{j}, $\hat{a}_{ij}$ is the corresponding element in the adjacency matrix, $\mathbf{x}_j$ is the input features of node \textit{j}, $\mathcal{N}^{i}$ is the set of connected nodes for node \textit{i}, and $\mathbf{h}_i$ is the updated features of node \textit{i}.

\noindent \textbf{Pyramidal Channel-Squeezing-Fusion Block (PCSF):}
\\
We hypothesize that the relevancy between two joints depends on their proximity according to the skeletal structure. This aligns with Information Gain analysis \cite{zhang2022information}, which proves that information gain diminishes exponentially as the node distance increases. Furthermore, clinical observation \cite{fahn2003description} suggests that PD patients typically experience PT on only one side of the upper body. Therefore, the information relevancy from one arm to another should be small.

\begin{figure}
\centering
\includegraphics[width=\textwidth]{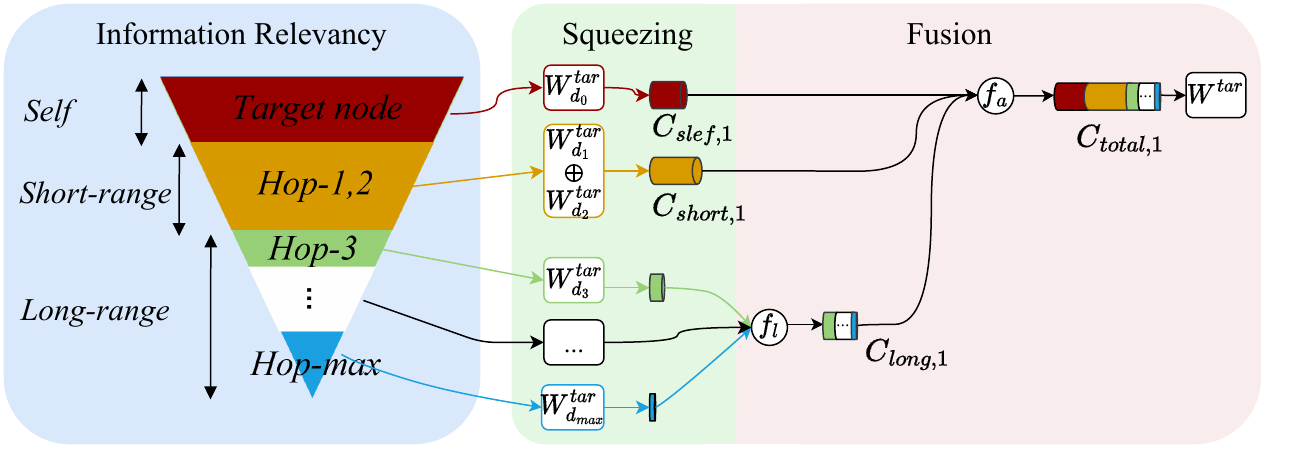}
\caption{The proposed Pyramidal Channel-Squeezing-Fusion architectures.}
\vspace{-5mm}
\label{fig1}
\end{figure}

To realize the hypothesis, we propose a novel lightweight PCSF that better models the relevancy of joints from their neighbors, thereby enhancing the network performance. As shown in Fig. \ref{fig1}, the output target node $i$'s attention weight $W^{i}$ is obtained from the joint-wise weights $\{W^{i}_{d_0},...,W^{i}_{d_{max}}\}$ after the squeezing-and-fusion process, where $d_n$ is the shortest distance between the target node $i$ and the relevant node $n$, namely \textit{Hop-n}. The visualization of information relevancy in Fig.\ref{fig1} guides the squeezing ratio, such that our method overcomes the limitation of the GCN (Graph Convolutional Network)~\cite{kipf2017semisupervised} that each joint shares the same weight. 

\vspace{-3mm}
\paragraph{The Channel-Squeezing Block:} 
We propose following squeezing operations to enhance the learning of PT-specific relevant information while filtering noise, based on our hypothesis motivated by \cite{zhang2022information,fahn2003description}. We distinguish nodes in different graph-distance by defining \textit{hop-0} node to be the self node, \textit{Hop-1,2} nodes to be the short-range nodes and \textit{Hop-3},...,\textit{Hop-max} to be the long-range nodes. Suppose the node \textit{i} is the target node, and the node \textit{j} is the relevant node of \textit{i}, then node \textit{j}'s output channel size is formulated by Eq.\ref{sq}:
\vspace{-1.5mm}
\begin{equation}
C_{out,j} = \begin{cases}
         C_{in}, & \lvert j-i\rvert = 0 \quad,\\
         p C_{in}, & 0<\lvert j-i\rvert \leq 2 \quad,\\
         q^{\lvert r-i\rvert}C_{in}, & \lvert j-i\rvert > 2.
        \end{cases}
    \label{sq}
\end{equation}
where \textit{p}, \textit{q} are channel-squeezing ratios for \textit{Hop-1,2} nodes and \textit{Hop-3,..., max} nodes, respectively. $p,q\in[0,1]$ and $p\gg q$. $C_{out,j}$ is the output channel size of node \textit{j}. $\lvert \cdot \rvert$ denotes the graph distance between nodes.

\vspace{-3mm}
\paragraph{The Channel-Fusion Block:} To hierarchically combine the different range information of the target node \textit{i}, we fuse the long-range features by $f_l$, and fuse all features by $f_a$:
\begin{equation}
    \mathbf{h_i}= {f_a} [\mathbf{h_{self}},\mathbf{h_{short}},f_{l}(\mathbf{h_{long,k}})]\mathbf{W^i}
    \label{para}
\end{equation}
where $\mathbf{h_{long,k}}$ is the feature of the long-range node \textit{k}, $\mathbf{h_{short}}$ and $\mathbf{h_{self}}$ are features of short-range nodes and self-node, respectively. $\mathbf{W^i}$ is the final weight matrix of target node \textit{i}.  

\vspace{-3mm}
\subsubsection{Implementation Details:}
\vspace{-1mm}
As depicted in Fig. \ref{net}, our network employs two GNN blocks with output channel sizes of 64 and 128, respectively. Each block contains an LCN layer (Locally Connected Network~\cite{ci2020locally}), a batch normalization layer, a LeakyReLU layer with an alpha of 0.2, and a dropout layer with 0.2 rates. Following the two GNN blocks, we employ a PCSF block, a global average pooling layer, and a fully connected layer. We adapt Cross-Entropy loss in binary classification. To address the class imbalance in multiclass classification, we use the focal-loss~\cite{lin2017focal} instead. Our optimizer of choice is Adam. The best performance of the PT binary classification task is achieved by a learning rate of 0.01 (decays by 0.1 ), a batch size of 8, and a dropout of 0.2, at 500 epochs. 

\vspace{-3mm}
\section{Experiments}
\vspace{-1mm}
To assess the efficacy of our proposed method, we conducted validation testing on two separate evaluation exams: the PT classification exam and the tremor rating estimation exam. We carried out our experiments using a Ubuntu 18.04 PC with an NVIDIA 3080. The GPU memory usage for training was minimal, averaging just 1.46 gigabytes. The training process for the TIM-TREMOR dataset took approximately ten hours for the PT classification task, and twelve hours for the tremor rating estimation exam. They include the processes of EVM and extraction of human pose features from RGB videos. In terms of real-time application, the PT classification or tremor rating estimation of a 33-second video with 1000 frames only took around 48 seconds each, which is a feasible time for interactive diagnosis.

\textbf{The Dataset:}
We test our system using the TIM-TREMOR dataset~\cite{pintea2018hand}, which is an open dataset consisting of 910 videos of 55 individuals performing 21 tasks. The videos are 18-112s long. There are 572 videos depicting various forms of tremors, including 105 for Parkinsonian Tremor (PT), 182 for Essential Tremor (ET), 88 for Functional Tremor (FT), and 197 for Dystonic Tremor (DT). An additional 60 videos (NT) were recorded without convincing tremors during the assessment. The test 278 videos have inconclusive tremor classification results and have been labeled as "Other." For the tremor rating labels, eight levels from level 0 to 7 are assigned to the individual's left and right hands, evaluated by Bain and Findley Tremor Clinical Rating Scale~\cite{Bain1993rating}. To ensure that there is only one label per video and preserve the characteristics of the video, we combine the labels for individual left and right hands by taking the maximum value of both hands.

\textbf{Setup:}
We eliminate inconsistent videos to minimize data noise, specifically videos that only capture motion tasks for a limited number of participants. For the tremor-type classification task only, we remove the videos with uncertain tremor-type labels of ``other.'' Next, each video is clipped into 100-frame samples, and the number of samples is determined by the duration of the consecutive frames in which the participant was visible and not obscured. Each sample was assigned the label of the original video and treated as an individual sample. We use a voting system to obtain the video-level classification results, which increases the system's robustness and augments the training sample size~\cite{lu2021quantifying}. We evaluate our proposed system through individual-based leave-one-out cross-validation. It means each subclip for a single individual is used for testing and excluded from the training set for each iteration. The subclips for each individual are never separated by the training or testing set. The total number of leave-one-out cross-validations are 39 and 55 for tremor-type classification and tremor-rating estimation, respectively.


\textbf{Evaluation Metrics:} We report the mean values calculated among all leave-one-out cross-validations with the following metrics: accuracy (AC), sensitivity (SE), specificity (SP), and F1-Score for the binary classification; AC, macro-averaged F1-score, SE and SP for the multiclass classification. 

\begin{table}
\footnotesize
\centering
\begin{tabular}{ccccccccc}
\hline
 &
\multicolumn{4}{c}{Binary Classification} &
\multicolumn{4}{c}{Multiclass Classification} 
\\
\hline
Method  &
AC & 
SE &
SP &
F1 &
AC & 
SE &
SP &
F1 
\\
\hline
ST-GCN\cite{yan2018spatial}  &
 84.6  &
 71.4  &
 87.5  &
 62.5  &
64.1&
64.8&
90.7&
64.1  
\\
CNN-Conv1D &
 76.9  &
 57.1  &
 81.3  &
 47.1&
56.4&
54.2&
88.3&
53.1 
\\
Decision Tree &
69.2&
57.1&
71.9&
40.0&
51.3&
49.4&
87.6&
36.7   
\\
SVM~\cite{wang2021hand}&
64.1&
57.1&
65.6&
36.4&
46.2&
44.6&
86.2&
44.3  
\\
\hline
Ours - full &
\textbf{92.3}&
\textbf{85.7}&
\textbf{93.8}&
\textbf{80.0}&
\textbf{71.8}&
\textbf{71.3}&
\textbf{92.5}&
\textbf{72.5}
\\
w/o PCSF &
87.2&
85.7&
87.5&
70.6&
66.7 &
67.6 & 
91.4 &
66.7
\\
w/o Attention &
82.1&
71.4&
84.4&
58.8&
61.5 &
63.1 &
90.0 &
62.4   
\\
w/o Attention \& EVM &
79.5&
71.4&
81.3&
55.6&
59.0 &
59.1 &
89.5 &
58.5   
\\
\hline
\end{tabular}
{\caption{The comparisons on the tremor type classification task.}\label{tab1}
}
\vspace{-5mm}
\end{table}

\vspace{-3mm}
\subsection{Tremor Type Classification}
\begin{figure}[htbp]
\centering
\vspace{-3mm}
\begin{minipage}[t]{0.43\textwidth}
\centering
\includegraphics[width=\textwidth]{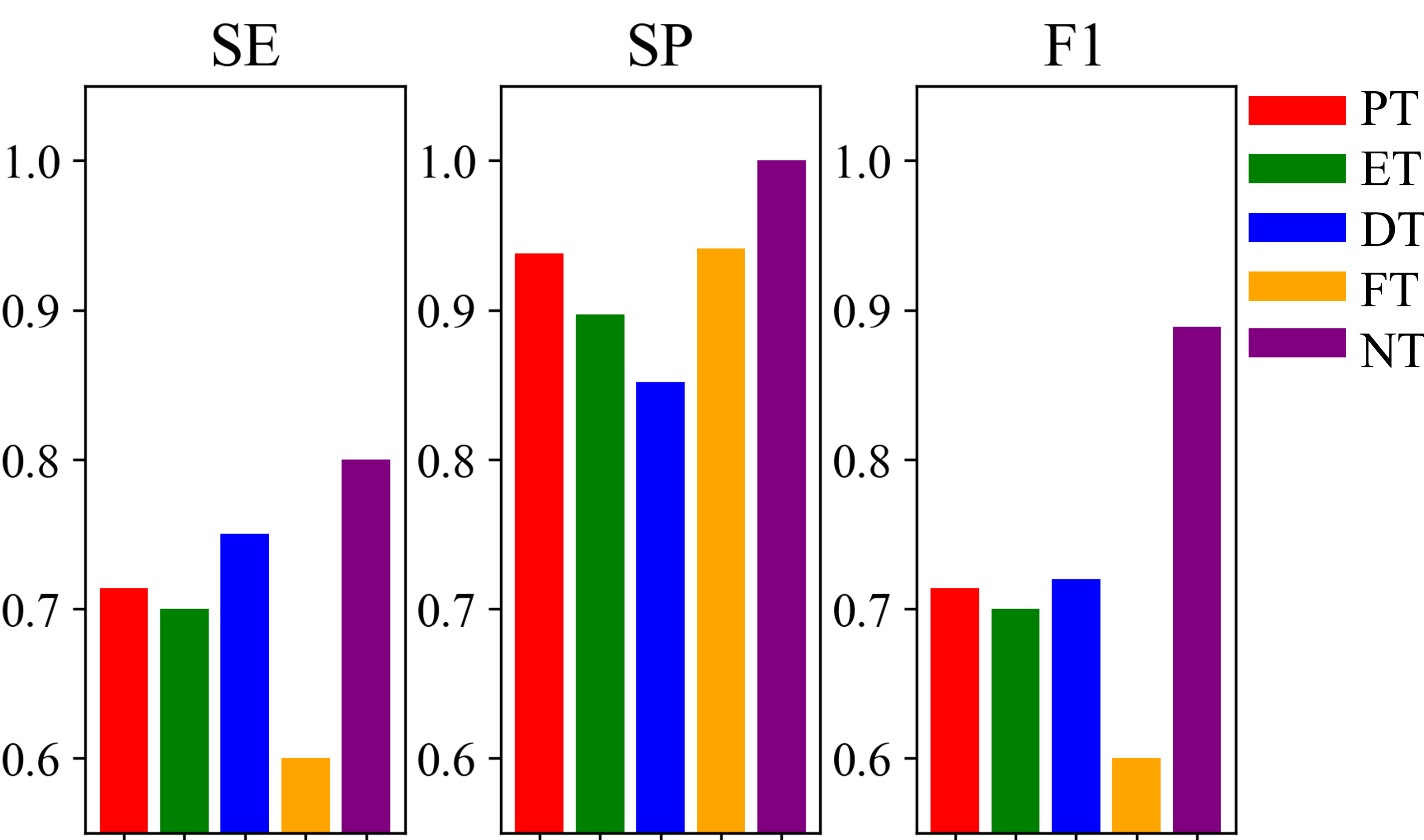}
\caption{Per-class multiclass tremor type classification results.}
\vspace{-5mm}
\label{fig3}
\end{minipage}
\begin{minipage}[t]{0.55\textwidth}
\centering
\includegraphics[width=\textwidth]{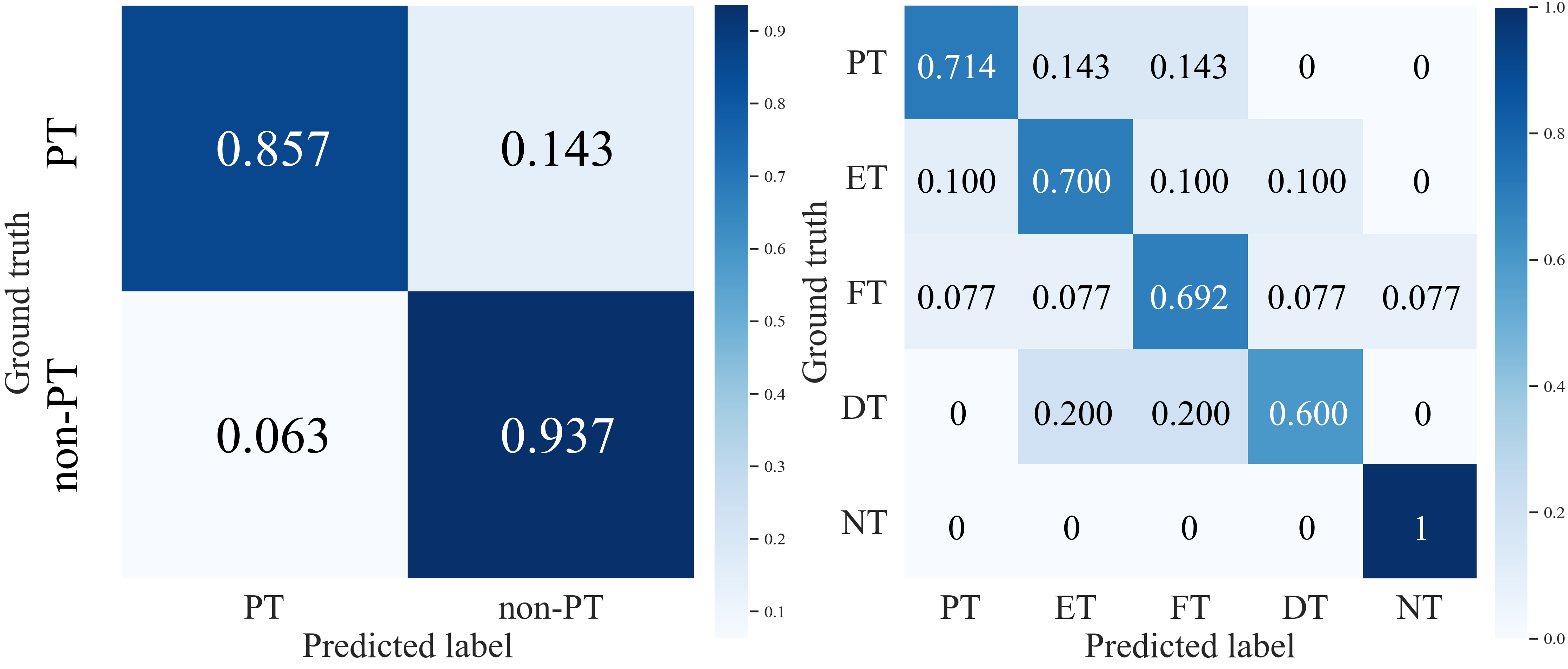}
\caption{Confusion matrices for PT classifications: (Left) binary; (Right) multiclass.}
\vspace{-5mm}
\label{fig4}
\end{minipage}
\end{figure}

\vspace{-1mm}
For this experiment, we first evaluate our system on the binary classification that distinguishes PT labels from non-PT labels, and achieve  91.3\% accuracy and 80.0\% F1-score. In addition, we validate our method on a more complex multiclass classification task for classifying five types of tremors (PT, ET,  DT, FT, and NT). Our final system's per-class tremor type multiclass classification performance is shown in Fig.~\ref{fig3}. It shows a fairly balanced performance on classifying PT, ET, DT, and NT, while FT has a lower SE and F1 score, which may be caused by the smallest number of samples in this class. Moreover, the corresponding confusion matrices of the two tasks are displayed in Fig.~\ref{fig4}.

\textbf{Comparison with Baseline Methods}: As this paper is the first work that provides the individual-level evaluation results, we implemented the following video-based PT classification baselines to evaluate the effectiveness of our system: (i) ST-GCN~\cite{yan2018spatial}: a spatial-temporal GCNs for human pose data classification; (ii) CNN with 1D convolutional layers (CNN-Conv1D)~\cite{wang2021hand}; (iii) Decision Tree (DT); (iv) Support Vector Machine (SVM)~\cite{wang2021hand}. Note that all baseline methods apply the same EVM and pose extraction design. The results of our proposed SPA-PTA and baselines are summarized in Table \ref{tab1}. 

The binary classification result shows that our full system consistently outperforms all other methods in all evaluation metrics. Our AC, SE, SP, and F1 achieve over 80\% on leave-one-out cross-validation, demonstrating the effectiveness and stability of our system in this binary classification task. It is noticeable that our system performs better with only spatial convolution instead of a deeper spatial-temporal convolution design like ST-GCN~\cite{yan2018spatial}. The outcome supports that the suggested PCSF block effectively enhances classification reliability and reduces the risk of overfitting in small datasets. 

While the full system is initially designed for binary classification, it presents effectiveness and generalizability in the multiclass classification task, surpassing existing methods. A small difference between AC, SE, and SP shows that our system performs consistently and effectively at identifying the positive samples and excluding the negative ones. The high macro-average SP exhibited trustworthy effectiveness in correctly recognizing individuals who have a specific type of tremor without wrongly recognizing it as other types of tremor.  

\textbf{Ablation Studies}:  We conduct an ablation analysis to assess the effectiveness of the EVM, PCSF block and the entire attention module. From the lower parts of Table \ref{tab1}, the positive effect of the PCSF block and attention module can be illustrated by the decrease in metrics when either the PCSF block or the entire attention module is removed in the two classification tasks. Also, we find that the basic GNN architecture without attention performs better than the CNN-Con1D model for both classification tasks. It highlights the efficacy of learning human pose features in the graph domain as opposed to the Euclidean domain. Besides, the variant of ``ours without attention'' performs slightly better than ``ours without attention and EVM preprocessing'', indicating that the use of EVM could effectively enhance tremors.

\begin{figure}
\includegraphics[width=\textwidth]{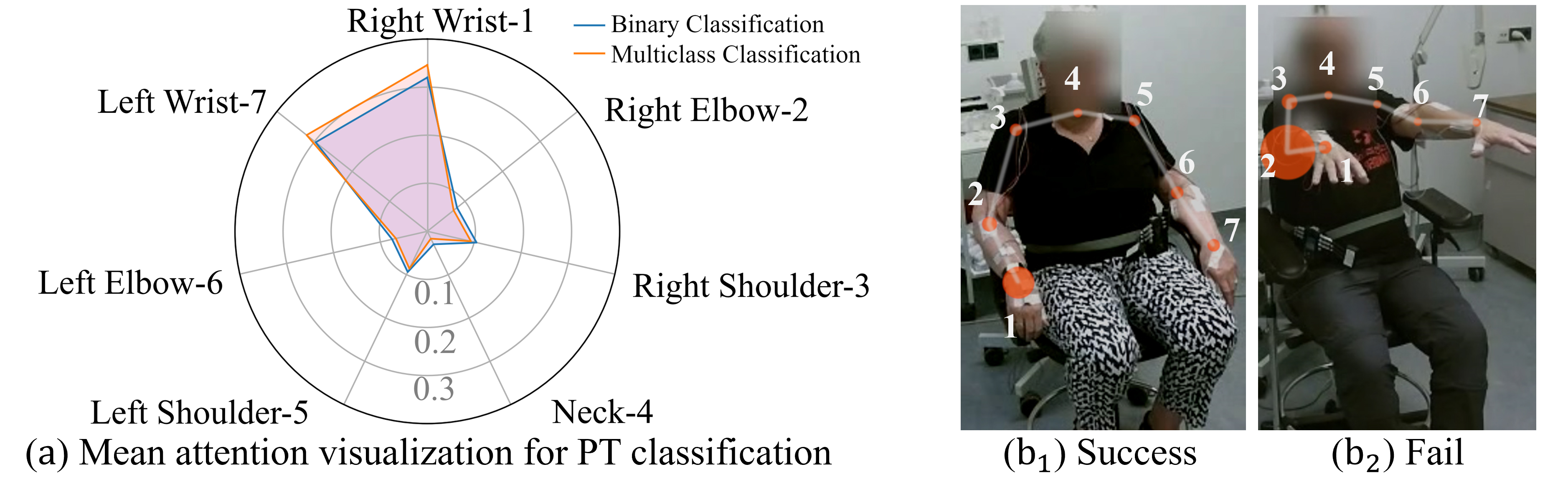}

\caption{(a) The average skeleton joints attention across all cross-validations in the PT classification experiment; (b) The attention visualization at a (b$_1$) successfully classified frame, and (b$_2$) unsuccessful classified frame. The joint labels in (b) correspond to (a); } \label{fig5}
\vspace{-5mm}
\end{figure}

\textbf{Model Interpretation:} We present the visualization for the average attention value of each body keypoint in Fig.~\ref{fig5}a. It is interpreted as the importance level our system considers during the classification process. Our analysis reveals that the attention value is significantly highest on the `Right Wrist' and `Left Wrist', which suggests that our system prioritizes the wrists' movements during the task performance. Furthermore, the value associated with the `Neck' is significantly lower than other keypoints. It may be explained by the fact that the participants remained seated during the video recording, resulting in a minimal global variance of the neck joint throughout the experiment.

\vspace{-3mm}
\subsection{Tremor Rating Estimation}
\vspace{-1mm}
For this experiment, we train SPA-PTA with different tremor rating labels without further implementation (e.g., converting the classification layer to a regression layer)  to validate our system performance in the tremor rating estimation task. Since the data with tremor ratings 4 and above is insufficient for training via leave-one-out cross-validation (i.e., only 5 individuals out of 55), we validate our system on two different classification settings: (1) Classifying ratings [1,2,3] (2) Classifying ratings [1,2,3+]. The latter is generally a more challenging task since the imbalanced data of the ``3+''  rating brings bias compared to the former, which does not contain such data.

\vspace{-5mm}
\begin{figure}[htbp]
\centering
\begin{minipage}[t]{0.64\textwidth}
\centering
\includegraphics[width=\textwidth]{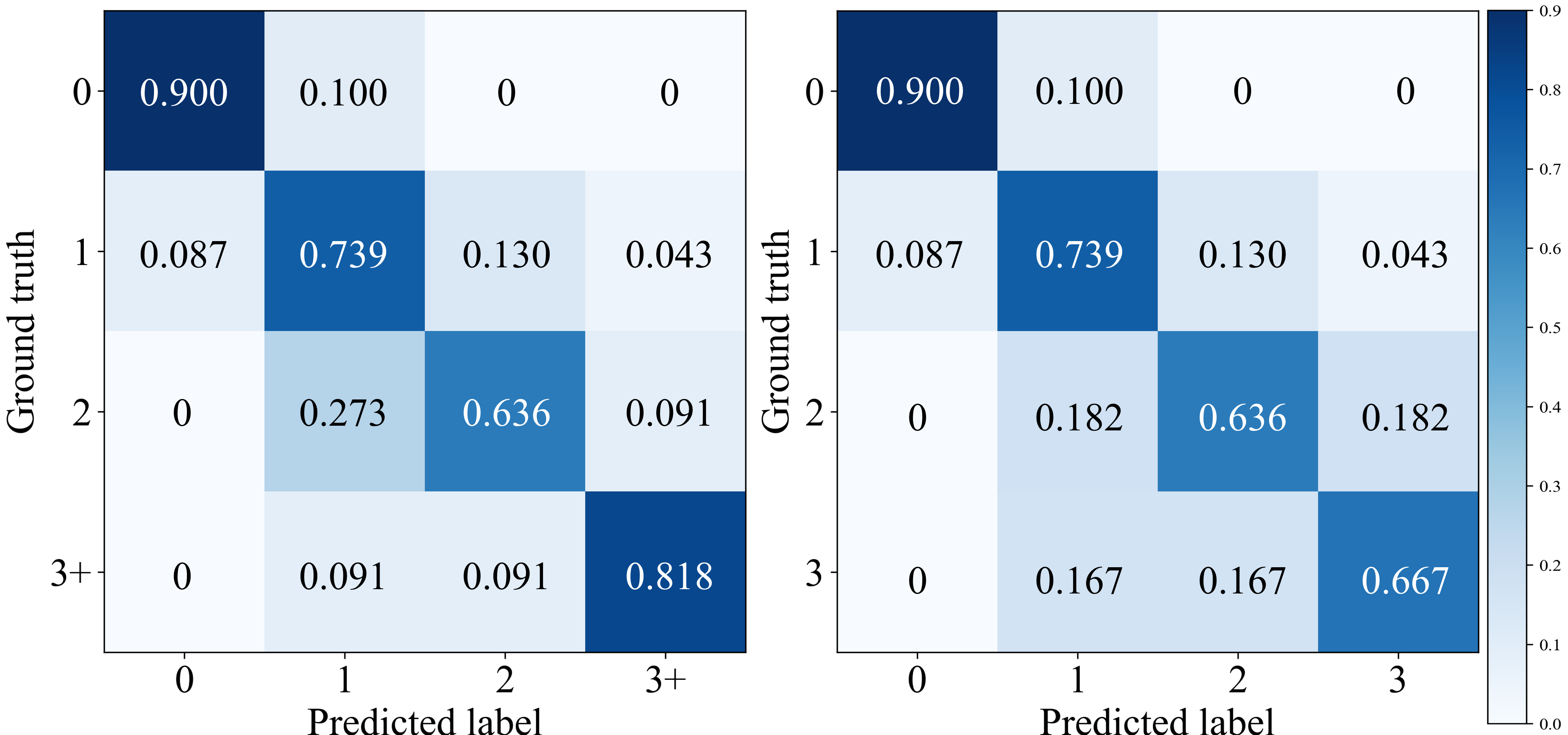}
\caption{Confusion matrices for tremor rating estimation: (Left) [1,2,3+]; (Right) [1,2,3].}
\label{fig6}
\end{minipage}
\begin{minipage}[t]{0.35\textwidth}
\centering
\includegraphics[width=\textwidth]{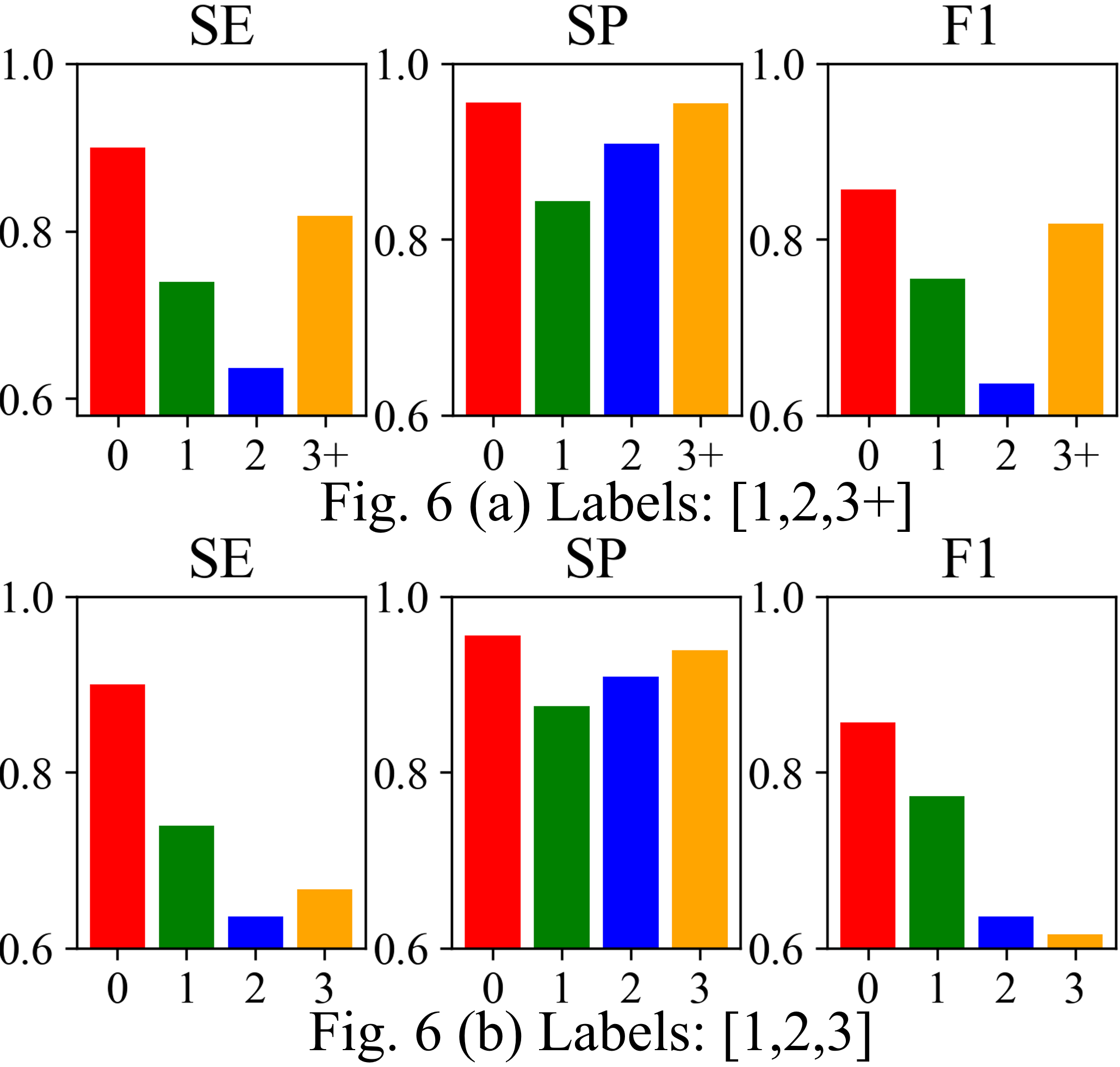}
\caption{Per-class tremor rating estimation results.}
\label{fig7}
\end{minipage}

\end{figure}

\vspace{-5mm}
\textbf{Comparison with Baseline Methods:} We compare our SPA-PTA to the same baselines in the tremor-type classification task as shown in Table.~\ref{tab4}. SPA-PTA significantly outperforms the baselines by achieving a macro-average AC of 76.4\%, SE of 77.3\%, SP of 91.6\%, and F1-score of 76.7\%. An interesting finding is that the machine learning-based method Decision Tree achieves similar performance to two deep learning-based baselines (i.e., ST-GCN and CNN-Conv1D). It may inform us to tackle the challenge of improving the deep learning models in a relatively small dataset. In addition, although our current model does not show strong robustness in the tremor rating estimation task, the ablation studies from the rows of "Ours" in Table~\ref{tab4} still demonstrate the effectiveness of our PCSF layer and the attention mechanism design. It shows the potential of improving our model and system performance with a more specific architecture design with a more extensive dataset.

\vspace{-0mm}
\begin{table}[h]
\centering
\footnotesize
\begin{tabular}{ccccccccc}
\hline
Classification labels&
\multicolumn{4}{c}{$[1,2,3+]$} &
\multicolumn{4}{c}{$[1,2,3]$} 
\\
\hline
Method  &
AC & 
SE &
SP &
F1 &
AC & 
SE &
SP &
F1 
\\
\hline
ST-GCN\cite{yan2018spatial}  &
67.3  &
68.1  &
89.0  &
66.5  &
68.0&
67.7&
90.5&
65.7  
\\
CNN-Conv1D &
60.0  &
59.8  &
86.5  &
58.7  &
60.0&
60.5&
87.9&
58.3 
\\
Decision Tree &
54.5&
55.3&
85.2&
54.6&
52.0&
53.0&
86.0&
51.3   
\\
SVM~\cite{wang2021hand}&
49.1&
41.1&
81.5&
43.8&
48.0&
49.5&
85.2&
47.1  
\\
\hline
Ours - full &
\textbf{76.4}&
\textbf{77.3}&
\textbf{91.6}&
\textbf{76.7}&
\textbf{74.0}&
\textbf{73.5}&
\textbf{92.0}&
\textbf{72.0}
\\
w/o PCSF &
70.9&
71.5&
89.7&
70.7&
70.0 &
68.6 & 
90.5 &
68.2
\\
w/o Attention &
65.5&
65.6&
88.2&
64.8&
66.0 &
65.2 &
89.5 &
63.9   
\\
w/o Attention \& EVM &
63.6&
64.8&
87.6&
63.3&
64.0 &
64.1 &
88.9 &
62.5   
\\
\hline
\end{tabular}
\caption{The comparisons on the tremor rating task. }
\label{tab4}
\vspace{-4mm}
\end{table}

\textbf{Ablation Studies}:  Consistent results at the bottom of Table~\ref{tab4} from the same ablation design as for the PT classification task validate the effectiveness of each system component.

\begin{figure}
\begin{minipage}[c]{0.6\textwidth}
\centering
\includegraphics[width=0.9\textwidth]{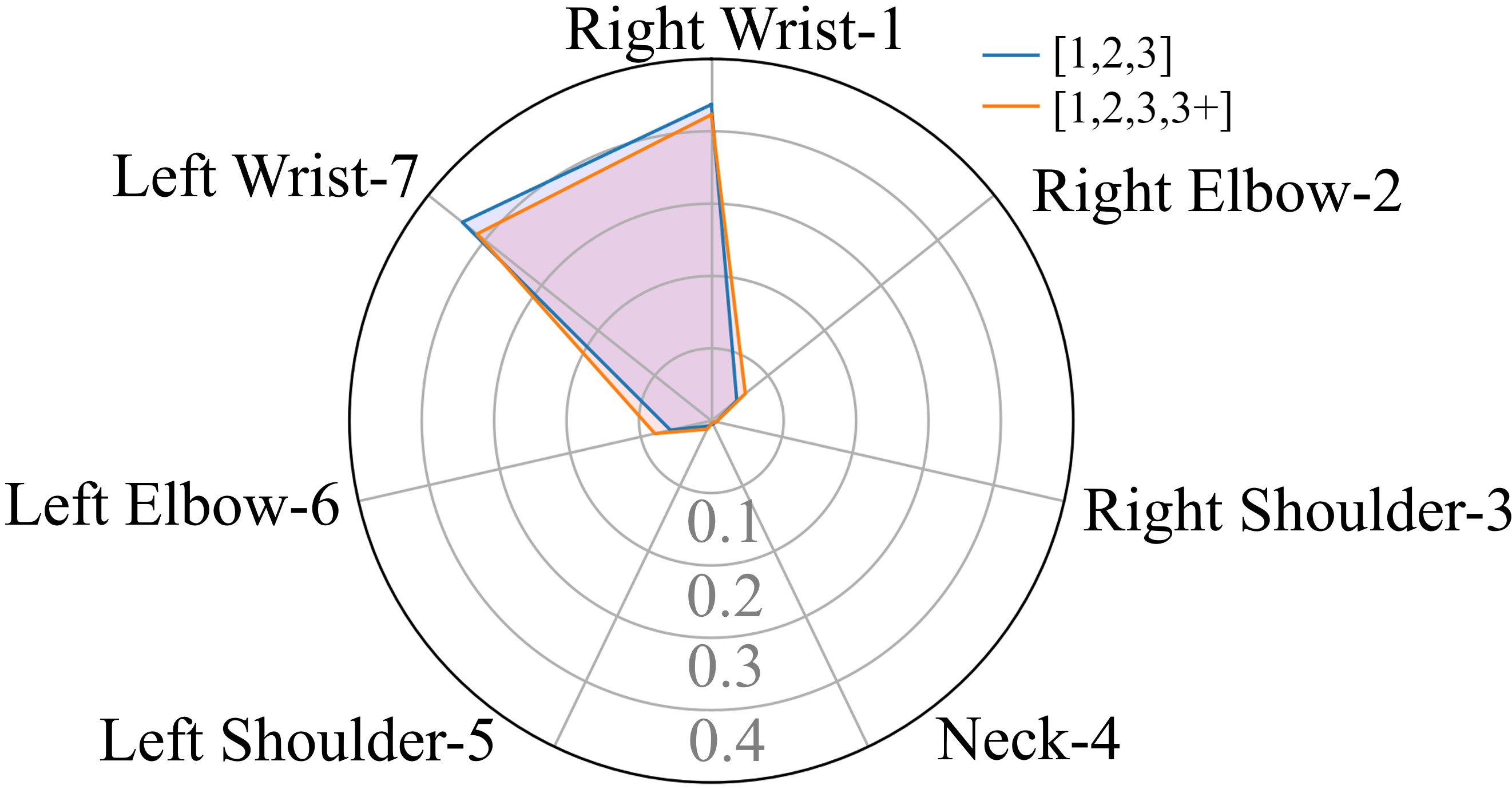}
\end{minipage}\hfill
\begin{minipage}[c]{0.37\textwidth}
\caption{The average skeleton joint attention across all cross-validations in tremor rating estimation task} 
\vspace{-8mm}
\label{fig8}
\end{minipage}
\end{figure}

\textbf{Model Interpretation}: 
We similarly visualize the average skeleton joints attention across all cross-validation sets in Fig.~\ref{fig8}. Two different data preprocessing approaches provide similar attention results, while the weights obtained by grouping [1,2,3] slightly more contribute to `Right Wrist' and `Left Wrist.' This may be due to the increased proportion of low tremor rating videos in this approach compared to grouping [1,2,3+]. In addition, we notice that the attention weight distribution of the tremor rating estimation exam is similar to that of the PT classification exam, while the former aggregates more attention on the `Right Wrist' and `Left Wrist' than other joints.

\subsection{Pose Estimation Evaluation}
To evaluate the effectiveness of AlphaPose and quantify the pose estimation error, we conduct the following experiments:\\
\textbf{Quantitative Comparison with Ground Truth Data}. To quantify the pose estimation error from different methods, we employ the Lagrangian hand-tremor frequency estimation method~\cite{pintea2018hand} to compare MAE (Mean Absolute Error) of the hand tremor frequencies estimated by AlphaPose and conventional OpenPose features~\cite{zhang2022miccai} with Ground Truth (GT) frequency obtained from accelerometer data. As suggested in~\cite{pintea2018hand}, tremor frequency calculated from reliable estimated pose features should be close to (i.e., ideally within 1 HZ difference) the GT accelerometer data frequency. The MAE from Table~\ref{tab_ao} indicates that AlphaPose consistently outperforms OpenPose on all listed tasks.

\begin{table}
\centering
\begin{tabular}{ccc}
\hline
Task&
AlphaPose&
OpenPose
\\
\hline
Rest&
\textbf{0.812}&
0.881
\\
Rest in supination&
\textbf{0.834}&
0.930
\\
2 Hz higher&
\textbf{0.605}&
0.635
\\
2 Hz lower&
\textbf{0.617}&
0.622
\\
Counting&
\textbf{0.729}&
0.790
\\
Finger tapping&
\textbf{0.576}&
0.687
\\
Playing piano&
\textbf{0.752}&
0.906
\\
Months backward&
\textbf{0.814}&
0.838
\\
Top top&
\textbf{0.786}&
0.823
\\
Thumbs up&
\textbf{0.844}&
0.960
\\
\hline
Average MAE&
\textbf{0.737}&
0.807\\
\hline
\end{tabular}
{\caption{MAE comparison between AlphaPose features and OpenPose on the top-10 best-performing tasks. Better performance with lower MAE is in bold.}\label{tab_ao}
}
\end{table}

\begin{figure}
\centering
\includegraphics[width=\textwidth]{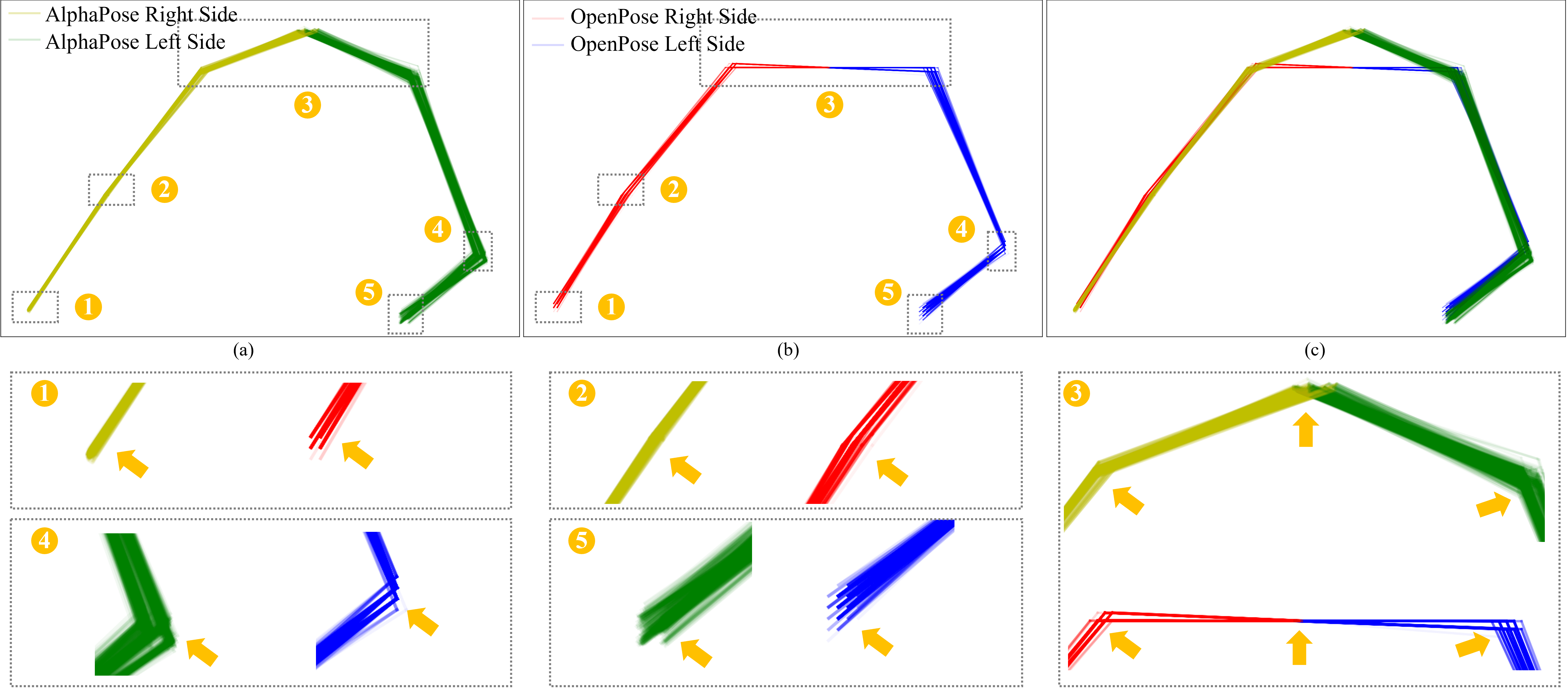}
\caption{The estimated pose comparison between AlphaPose and OpenPose for a sitting and resting PD patient with clinically identified PT on the left side of the body. (a), (b) and (c) are the estimated poses of an example video from AlphaPose, OpenPose, and both, respectively. Each colored line with 0.05 transparency represents the connection between joints estimated in each frame. Numbers 1 to 5 correspond to specific joints' local scaling for intuitive comparison. The raw video frames are referenced in Fig.~\ref{fig10}}.\label{fig9}
\end{figure}

\begin{figure}
\centering
\includegraphics[width=\textwidth]{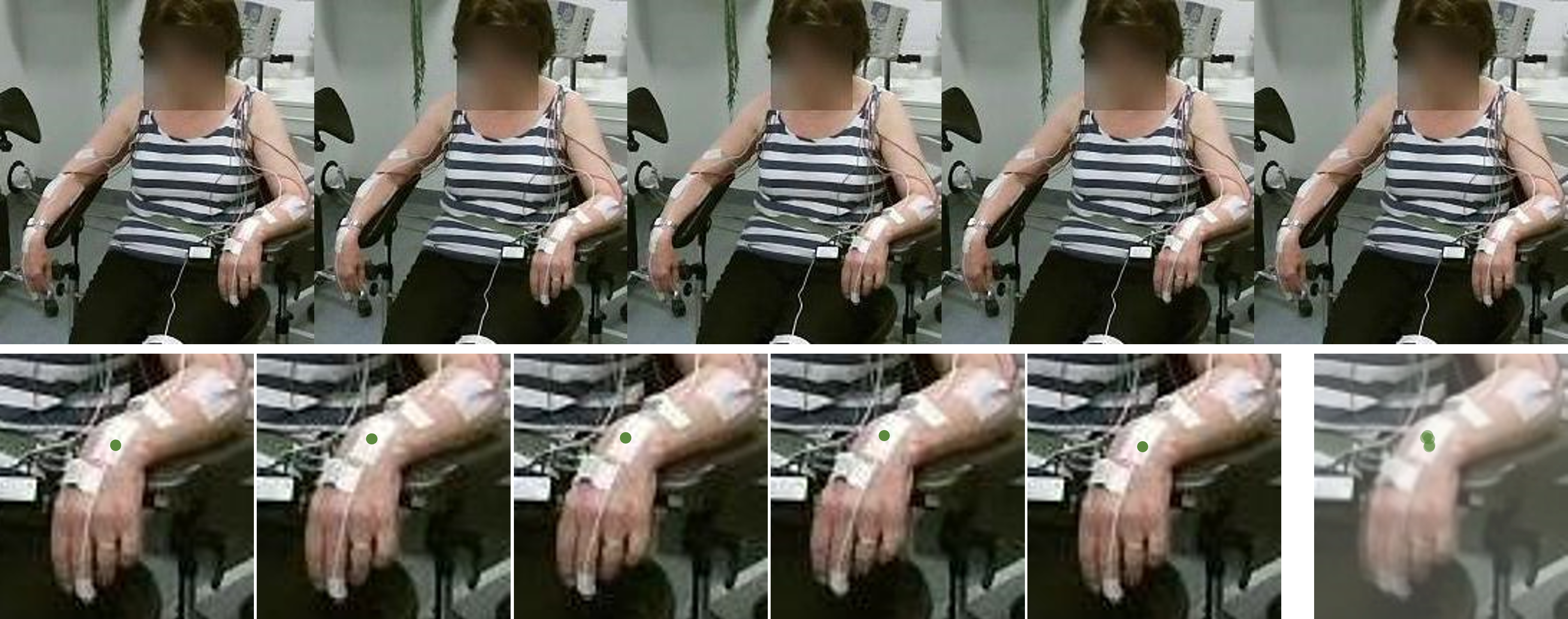}
\caption{The raw videos referenced in Fig.~\ref{fig9} consist of consecutive images captured at intervals of 5 frames, approximately every 0.167 seconds. The lower-right image is an aggregation of five transparent hand images, where the green dot shows the estimated trajectory of the left wrist joint during tremor.}
\label{fig10}
\vspace{-3mm}
\end{figure}

\textbf{Qualitative Pose Visualization and Comparison}. The visualizations in Fig.~\ref{fig9} and the reference video images in Figure~\ref{fig10} show that AlphaPose outperforms OpenPose in estimating joint positions. This is supported by the smoother trajectory lines of AlphaPose, which are depicted by the transparent colored lines. Sub-figures 1 to 5 in Fig.~\ref{fig9} demonstrate AlphaPose's ability to track joint movement fluidly. Specifically, in sub-figure 5, AlphaPose demonstrates a hand trajectory that aligns more closely with the anticipated tremor pattern, which contrasts with OpenPose's intermittent jumping trajectory. This consistency suggests that AlphaPose may be more reliable for tasks related to PT classification. Furthermore, on the patient's right side, particularly in sub-figures 1 and 2, AlphaPose yields more consistent and stable outcomes, reflecting the patient's condition of resting with observable tremors only in the left hand, as corroborated by Figure 10. Finally, the neck joint position of OpenPose is estimated by the mean point of both shoulders, which is less accurate than the estimated neck joint position of AlphaPose~\cite{alphapose}.\\

\vspace{-3mm}
\textbf{Classification Performance Comparison}. We compare the effectiveness of AlphaPose and OpenPose by evaluating their impacts on the system classification performance. Table~\ref{tab_avoc} demonstrates that using AlphaPose features results in a remarkable and consistent improvement over OpenPose features of approximately $1-3\%$ across the classification tasks except for the binary tremor-type classification. These results highlight the precision of AlphaPose in delivering better pose-based features for classification tasks.

\begin{table}
\centering
\begin{tabular}{ccccccccc}
\hline
Method  &
AC & 
SE &
SP &
F1 &
AC & 
SE &
SP &
F1 
\\
\hline
 &
\multicolumn{4}{c}{Tremor-type Binary } &
\multicolumn{4}{c}{Tremor-type Multiclass } 
\\
AlphaPose &
\textbf{92.3}&
\textbf{85.7}&
\textbf{93.8}&
\textbf{80.0}&
\textbf{71.8}&
\textbf{71.3}&
\textbf{92.5}&
\textbf{72.5}
\\
OpenPose &
\textbf{92.3}&
\textbf{85.7}&
\textbf{93.8}&
\textbf{80.0}&
69.2&
69.3&
92.0&
70.2
\\
\hline
 &
\multicolumn{4}{c}{Tremor-level [1,2,3+] } &
\multicolumn{4}{c}{Tremor-level [1,2,3] } 
\\

AlphaPose &
\textbf{76.4}&
\textbf{77.3}&
\textbf{91.6}&
\textbf{76.7}&
\textbf{74.0}&
\textbf{73.5}&
\textbf{92.0}&
\textbf{72.0}
\\
OpenPose &
72.7&
74.0&
90.4&
73.6&
72.0 &
72.5 &
90.4 &
70.1   
\\
\hline
\end{tabular}
{\caption{The comparisons on the influence of classification performance between AlphaPose and OpenPose.}\label{tab_avoc}
}
\vspace{-3mm}
\end{table}

In this study, we utilize the pre-trained AlphaPose model, opting not to retrain it due to the absence of GT 2D pose position annotations within our dataset. The robust generalization capability of the pre-trained AlphaPose model, as evidenced by its superior performance across multiple diverse and complex benchmark datasets~\cite{alphapose}, affirms its suitability for our task. In the future, we are interested in comparing the performance between pretrained and tremor-specific pose estimation models. This will entail the collection of the necessary GT data to train a model adept at detecting the subtle nuances characteristic of tremor movement patterns.

\section{Conclusion and Discussion}
\vspace{-1mm}
Our method effectively identifies PT in PD patients from consumer-grade videos. The validity of our proposed system on both PT classification and tremor severity estimation tasks demonstrates that our method is extensible in PT-related analysis. Our non-intrusive system only relies on consumer-grade videos as input, so it offers a potentially cost-effective solution for supporting the pre-diagnosis of PD in regions with inadequate medical resources. This work could also be used for remote PD supplementary assessment in special situations to reduce the stress of the healthcare system (e.g., the COVID-19 pandemic). Moreover, our system demonstrates the potential to automatically monitor PT symptoms during daily activities to support PD pre-diagnosis.

Our findings about PT analysis are preliminary, and the limited number of people with PT and the limited range of tremor levels included in this work may affect the generalizability of the results. One of our future directions is to evaluate our models using data collected from a larger and more diverse group of Parkinson's disease patients, covering a more balanced tremor-type distribution and a wider range of tremor severity ratings. Up-scaling the study is crucial for developing more robust models and for enhancing the overall applicability and validity of the framework we have presented. In addition, annotating the dataset based on PT severity estimation performance by different scales, such as the MDS-UPDRS3, by experienced raters will enable us to improve the robustness of our model in the future. Moreover, our current system performance is still challenged by pose estimation algorithm error, such as depicted in blue{Fig. \ref{fig5}b}. The attention of our system is incorrectly influenced by the inaccurate position detection of the right elbow and blurred right shoulder joints.
\\[0.1pt]





\vspace{-5mm}
\section*{Statements and Declarations}
\vspace{-1mm}

\small \noindent \textbf{Funding:} H. Shum received support from the EPSRC NortHFutures project (Ref: EP/X031012/1).
S. Del Din has received support from Innovative Medicines Initiative 2 Joint Undertaking (Ref: 820820  Mobilise-D, 853981 IDEA-FAST), NIHR Newcastle Biomedical Research Centre, Newcastle upon Tyne Hospitals NHS Foundation Trust, Cumbria Northumberland and Tyne and Wear NHS Foundation Trust.

\small \noindent \textbf{Conflict of interest} S. Del Din reports consultancy activity with Hoffmann-La Roche Ltd. outside of this study.

\small \noindent \textbf{Ethical approval} Approval of the TIM-TREMOR dataset was obtained from the University Leiden University Medical Center ethics committee. The procedures used in this study adhere to the tenets of the Declaration of Helsinki.

\small \noindent \textbf{Informed consent} Informed consent was obtained from all individual participants included in the study.
\vspace{-0mm}
\setlength{\bibsep}{1pt}

\end{document}